\documentclass{article}
\usepackage{ijcai23}

\usepackage{times}
\usepackage{soul}
\usepackage{url}
\usepackage[hidelinks]{hyperref}
\usepackage[utf8]{inputenc}
\usepackage[small]{caption}
\usepackage{graphicx}
\usepackage{amsmath}
\usepackage{amsthm}
\usepackage{booktabs}
\usepackage{algorithm}
\usepackage{algorithmic}
\usepackage[switch]{lineno}

\title{A Survey on Compositional Generalization in Applications}

\author{
Baihan Lin$^1$, Djallel Bouneffouf$^2$, Irina Rish$^3$\\
\affiliations
$^1$ Columbia University, New York, NY 10027, USA\\
$^2$ IBM TJ Watson Research Center, Yorktown Heights, NY 10598, USA\\
$^3$ Mila - Quebec AI Institute, Montreal, Quebec H2S 3H1, Canada\\
\emails
baihan.lin@columbia.edu, djallel.bouneffouf@ibm.com, irina.rish@mila.quebec
}

\begin{document}

\maketitle

\begin{abstract}
The field of compositional generalization is currently experiencing a renaissance in AI, as novel problem settings and algorithms motivated by various practical applications are being introduced, building on top of the classical compositional generalization problem. This article aims to provide a comprehensive review of top recent developments in multiple real-life applications of the compositional generalization. Specifically, we introduce a taxonomy of common applications and summarize the state-of-the-art for each of those domains. Furthermore, we identify important current trends and provide new perspectives pertaining to the future of this burgeoning field.
\end{abstract}

\section{Introduction}

The ability to make generalizations based on different actions or concepts, known as systematic compositionality, is crucial for human learning and understanding \cite{lake2019human}. It allows us to understand and learn new things, even when we have limited experience. In human daily life, we often encounter problems that require us to be able to make generalizations based on compositions of different actions or concepts. This ability is critical for how we are able to learn and understand new things, even when we have limited experience. However, while there have been significant advances in the language capabilities of machines, they still struggle with generalization and require large amounts of training data. These machine learning techniques, like neural networks, have been criticized in the past for lacking systematic compositionality.

In many previous works, such as \cite{chomsky1957logical,fodor1975language,montague1970pragmatics}, the authors study the extent and nature of human compositional learning abilities through artificial tasks that involve mapping instructions made up of pseudowords to responses made up of colored circles. These tasks are designed to minimize reliance on knowledge of a specific language and to allow for direct comparison between humans and recurrent neural networks. The tasks also present a novel challenge for both humans and machines, as they must perform "few-shot learning" using only a handful of training examples. The results of these tasks will provide insight into the extent of human compositional learning abilities and the ways in which contemporary neural networks fail at these tasks.

To evaluate compositional learning, \cite{lake2018generalization} introduced the SCAN dataset for learning instructions such as walk twice and jump right,'' which were built compositionally from a set of primitive instructions (e.g., ``run'' and ``walk''), modifiers (twice'' or ``right''), and conjunctions (``and'' or ``after''). The authors found that modern recurrent neural networks can learn how to run'' and to run twice'' when both of these instructions occur in the training phase, yet fail to generalize to the meaning of ``jump twice'' when ``jump'' but not ``jump twice'' is included in the training data.

    In the past several years,  an increasing number of works has been focused on improving compositional generalization of neural networks. In particular, a lot of attention was devoted to learning   disentangled representations - i.e. a set of different factors in  some low-dimensional representation that jointly generate the observations, and multiple methods aiming to learn disentangled representation have been proposed recently \cite{higgins2017beta,kim2018disentangling,chen2018isolating,mathieu2019disentangling} summarized in \cite{locatello2019challenging}. A recent work by {\cite{xu2022compositional} compares such approaches versus another method, known as emergent language learning \cite{havrylov2017emergence}.  

We will now provide an overview including various applications of the compositional generalization framework, both in real-life problem setting arising in multiple practical domains (healthcare, computer network routing, finance, and beyond), as well as in computer science and machine-learning in particular, where compositional generalization approaches can help important algorithmic in unsupervised learning, supervised learning and reinforcement learning. These applications can help to address the challenges that machine learning techniques currently face in terms of generalization and require less amount of data. Additionally, this framework can be used to improve the understanding of human compositional learning abilities and the ways in which contemporary neural networks fail at these tasks.

\section{ Compositional Generalization }

\subsection{Formal definition} 

Montague, in his work \cite{montague1970pragmatics}, proposed a clear way to formally capture the principle of compositionality in language. The main idea is that compositionality requires there to be a relationship between the expressions of a language and the meanings of those expressions. To achieve this, Montague proposed a syntactic algebra, which is a partial algebra that consists of a set of expressions (simple and complex) and a number of operations (syntactic rules) defined on it \cite{szabo2004compositionality}. These operations always apply to a fixed number of expressions and produce a single expression, and they may not be defined for certain expressions.

The syntactic algebra is interpreted through a meaning assignment function, which maps from the set of expressions to the set of available meanings for those expressions. This meaning assignment function is said to be compositional if there is a partial function that maps the meanings of the expressions used to form a new expression through the application of a syntactic rule to the meaning of the resulting expression. In other words, there is a partial function that maps the meanings of the expressions used to form a new expression through the application of a syntactic rule to the meaning of the resulting expression. Formally, given an operation $F$, a $k$-ary syntactic operation on $E$, the meaning assignment function $m$ is $F$-compositional if there is a $k$-ary partial function $G$ on $M$ such that whenever $F(e_1, ..., e_k)$ is defined, $m(F(e_1, ..., e_k)) = G(m(e_1), ..., m(e_k))$. It is important to note that for $m$ to be compositional overall, it must be $F$-compositional for every syntactic operation in $E$. When $m$ is compositional, there exists a semantic algebra $M = <M, (G_\gamma)_{\gamma \in \Gamma}>$ on $M$, which is a homomorphism between $E$ and $M$ as shown in \cite{westerstaahl1998mathematical}. 

Montague introduces the idea of global and collective compositionality, which are extensions of the idea of local compositionality. Global compositionality states that two expressions are considered equivalent if they are produced by applying the same syntactic operation to lists of expressions such that corresponding members of the lists are either simple and synonymous or complex and globally equivalent. This is defined recursively, where two expressions are considered 1-global equivalents if they are synonymous simple expressions. Two expressions are considered $n$-global equivalents if for some natural number $k$, there is a $k$-ary $F$ in $E$, and there are expressions $e_1, ..., e_k, e'_1, ..., e'_k$ in $E$ such that $e = F(e_1, ..., e_k)$ and $e' = F(e'_1, ..., e'_k)$, and for every $1 \leq i \leq k$ there is a $1 \leq j < n$ such that $e_i$ and $e'_i$ are $j$-global equivalents. The meaning assignment function is considered globally compositional if globally equivalent pairs of expressions are all synonymous.

Collective compositionality is similar to global compositionality, but with one additional requirement. In the recursive step, we require not only that $e_i$ and $e'_i$ be $j$-collective equivalents, but also that the same semantic relationships should hold among $e_1, ..., e_k$ and among $e'_1, ..., e'_k$. This allows for the possibility that "Cicero is Cicero" is not collectively equivalent to "Cicero is Tully", even though they have the same structure and their proper constituents are all collectively equivalent. The meaning assignment function is considered collectively compositional if collectively equivalent pairs of expressions are all synonymous.

As pointed by \cite{szabo2004compositionality}, for more details on variants, and formal results on this topic, one can refer to \cite{janssen1997compositionality,hodges2001formal,pagin2010compositionality}. For generalizations that cover languages with various sorts of context-dependence, one can refer to \cite{pagin2005compositionality,pagin2007content,westerstaahl2012compositionality}.

\subsection{Productivity}
This idea dates back to Frege (1980) believed that our ability to understand sentences we have never heard before relies on the ability to construct the meaning of a sentence out of parts that correspond to words \cite{frege1980foundations}, which established the argument in favor of compositionality based on productivity. This argument is an inference to the best explanation, which can be expanded without assuming that meanings are Fregean senses \cite{szabo2004compositionality}.

The argument from productivity states that since competent speakers can understand a complex expression that they have never encountered before, it must be that they have a knowledge (perhaps implicit) which allows them to figure out the meaning of the expression without any additional information. This knowledge must be based on the structure of the expression and the individual meanings of its simple constituents. To support this claim, philosophers often appeal to the idea of unboundedness, which states that although we are finite beings, we have the capacity to understand an infinite number of complex expressions. Additionally, the fact that natural languages are learnable is also used to argue for compositionality. This is not an independent argument, as the reason it is remarkable that we can learn a natural language is that once we have learned it, our understanding is productive. This argument, however, is in favor of the $(C)$—global distributive language-bound compositionality of meaning, which cannot establish $(C_{ref}), (C_{local})$, or $(C_{cross})$. 

The argument from productivity, which states that our ability to understand complex expressions that we have never encountered before is based on our knowledge of the structure and meanings of the expression's simple constituents, can be criticized for not being able to establish a universal claim. Just because we tend to understand complex expressions we've never heard before, it doesn't mean that we will understand all complex expressions we come across. Even if in general we understand complex expressions based on the knowledge of their structure and meanings, there may be exceptions to this rule. Therefore, it's important to note that general considerations of productivity cannot rule out isolated exceptions to compositionality. However, if we limit our scope to proving that natural languages by and large obey global distributive language-bound compositionality of meaning, the argument from productivity is a strong one.

\subsection{Systematicity} 
As well defined in \cite{szabo2004compositionality}, systematicity refers to the systematic and regular way in which the meaning of a complex expression is determined by the meanings of its parts and the way they are syntactically combined. This property is known as compositionality. For example, consider the sentence ``The apple is on the tree.'' The meaning of this sentence is determined by the meanings of the words ``apple,'' is,'' on,'' and ``tree,'' and the way they are put together according to the rules of syntax. The meaning of the sentence is not a property of the words in isolation, but rather depends on the way they are combined. Systematicity is a key feature of human language and is one of the things that sets it apart from other forms of communication. It allows us to generate an infinite number of new sentences using a finite set of words and rules, and to understand the meaning of sentences we have never heard before.

Another argument in favor of compositionality is based on systematicity, the fact that there are definite and predictable patterns among the sentences we understand. For example, as an inference to the best explanation, understanding ``The car is parked in the garage'' entails understanding ``The garage is where the car is parked'' and vice versa. Or more formally, understanding a complex expression $E$ and $e'$ built up through the syntactic operation $F$ from constituents $e_1,...,e_n$ and $e'_1,...,e'_n$ respectively, can also imply understanding any other meaningful complex expression $e''$ built up through $F$ from expressions among $e_1,...,e_n,e'_1,...,e'_n$. So, it must be that anyone who knows what $E$ and $e'$ mean is in the position to figure out, without any additional information, what $e''$ means. If this is so, the meaning of $E$ and $e'$ must jointly determine the meaning of $e''$. But the only plausible way this could be true is if the meaning of $E$ determines $F$ and the meanings of $e_1,...,e_n,$, the meaning of $e'$ determines $F$ and the meanings of $e'_1,...,e'_n$, and $F$ and the meanings of $e_1,...,e_n,e'_1,...,e'_n,$ determine the meaning of $e''$.

\section{Real-Life Applications of Compositional Generalization }

As a general mathematical framework, the compositional generalization setting addresses the challenges associated with the presence of uncertainty in sequential decision-making. This type of uncertainty has a complex interplay with  the exploration versus exploitation dilemma, and therefore provides a natural formalism for most real-life  online decision-making problems.

\subsection{Mathematics}

In mathematics, compositional generalization refers to the ability of a model to understand and apply mathematical operations to novel combinations of mathematical elements that it has not encountered during training. For example, if a model has been trained to add numbers, it should be able to apply this operation to new sets of numbers it has never seen before, rather than only the specific numbers it was trained on. This ability is important for developing models that can solve mathematical problems and perform mathematical reasoning in a wide range of contexts. In \cite{lan2022improving}, the authors focus on analyzing the concept of compositional generalization in the context of math word problems (MWPs, Figure \ref{fig:exp_math}). The authors point out that while there has been a lot of research on compositional generalization in language and problem solving, there is limited discussion on this topic in relation to MWPs. To address this gap, the authors propose a novel method for studying compositional generalization in MWPs. This method involves creating compositional splits from existing MWP datasets, as well as synthesizing new data to isolate the effect of compositions. Additionally, the authors propose an iterative data augmentation technique that incorporates diverse compositional variations into training data and can be used in conjunction with existing MWP methods. The authors evaluate a set of different methods and find that all of them experience a significant loss in performance on the evaluated datasets. However, they also find that their data augmentation method significantly improves the compositional generalization of general MWP methods.

\begin{figure}[h!]
\centering
\includegraphics[width=\linewidth]{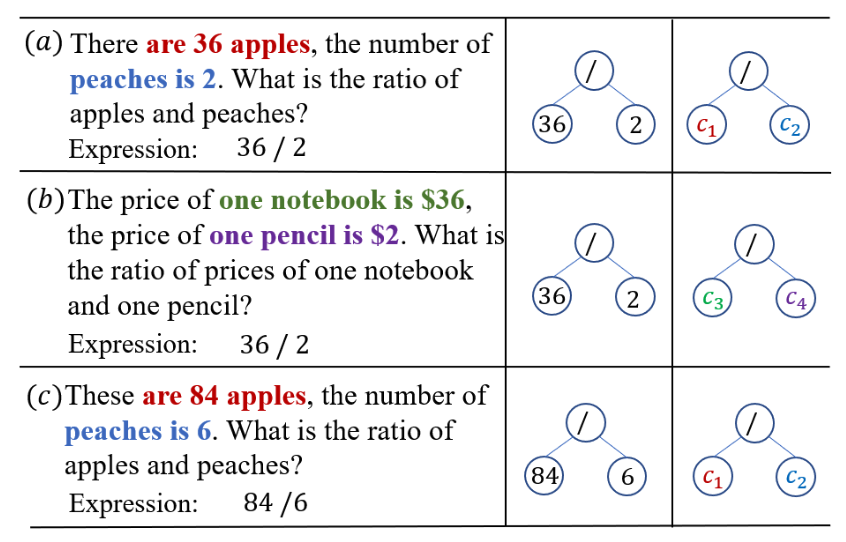}
\caption{Example of a Math Word Problem (MWP) and the compositional generalization of the expressions and contexts \protect\cite{lan2022improving}.
}\label{fig:exp_math}
\end{figure}

\subsection{Control systems}
In control systems, compositional generalization refers to the ability of a model to understand and control different entities in an environment, such as robots or machines, based on their relationships and interactions. This ability is important for developing models that can adapt to new environments, such as a new manufacturing facility or a new type of machine, and perform tasks in those environments with minimal additional training. In \cite{zhou2022policy}, the authors propose a new framework for learning goal-conditioned policies in control, robotics, and planning tasks. The framework is designed to handle tasks that have complex, compositional structures that can vary in the number of entities involved (e.g. Figure \ref{fig:exp_robot}). They introduce architectures such as Deep Sets and Self Attention which are designed to leverage the compositional structure of the tasks. The policies that are generated from these architectures are flexible and can be trained without the need for action primitives. The researchers evaluate their framework on a suite of simulated robot manipulation tasks, and find that the policies generated achieve higher success rates with less data, and are able to generalize to different numbers of entities and compositions than what was seen in the training data.
\begin{figure}[h!]
\centering
    \includegraphics[width=\linewidth]{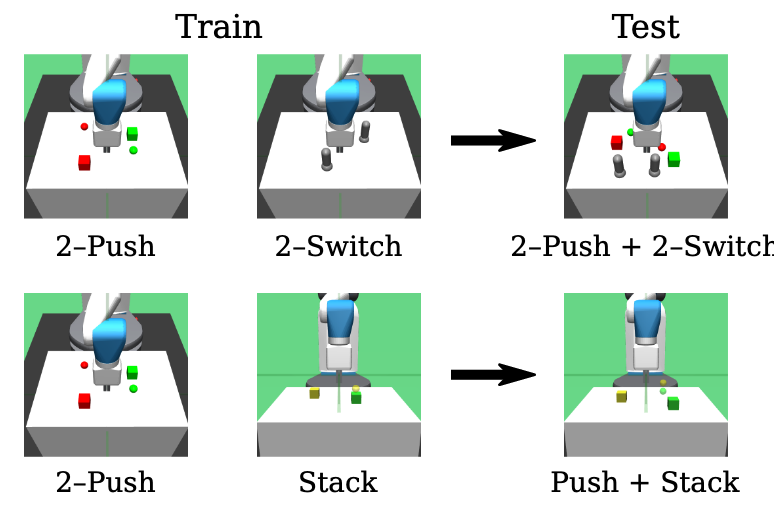}
\caption{Example of the compositional generalization in learning the robotic control (e.g skill stitching) \protect\cite{zhou2022policy}.
}\label{fig:exp_robot}
\end{figure}

\subsubsection{Reinforcement learning}
In reinforcement learning, compositional generalization refers to the ability of an agent to learn and generalize to new combinations of tasks, such as variations in number of entities or compositions of goals, in an environment. This ability is important for developing models that can perform complex tasks and adapt to new situations with minimal additional training. In \cite{mambelli2022compositional}, the authors focus on developing models that can effectively learn manipulation tasks in multi-object settings (e.g. Figure \ref{fig:exp_rl}) and perform well even when the number of objects changes. They examine the task of moving a specific cube out of a set to a goal position and find that current approaches, which rely on attention and graph neural networks, do not generalize well to different number of objects. To address this issue, the authors propose a new module that incorporates relational inductive biases, which allows the model to perform well and generalize to new scenarios with different number of objects. The proposed approach surpasses previous methods in performance and scales linearly with the number of objects, making it more efficient for extrapolation and generalization. In another work by \cite{zhao2022toward}, the authors approach the problem of object-oriented reinforcement learning by formalizing it with an algebraic method and studying how a world model can achieve it. They introduce a test environment called Object Library and use it to measure the generalization ability of different methods. They analyze several existing methods and find that they have limited or no compositional generalization ability. To address this, they propose a new approach called Homomorphic Object-oriented World Model (HOWM) that achieves soft, but more efficient compositional generalization.
\begin{figure}[h!]
\centering
    \includegraphics[width=\linewidth]{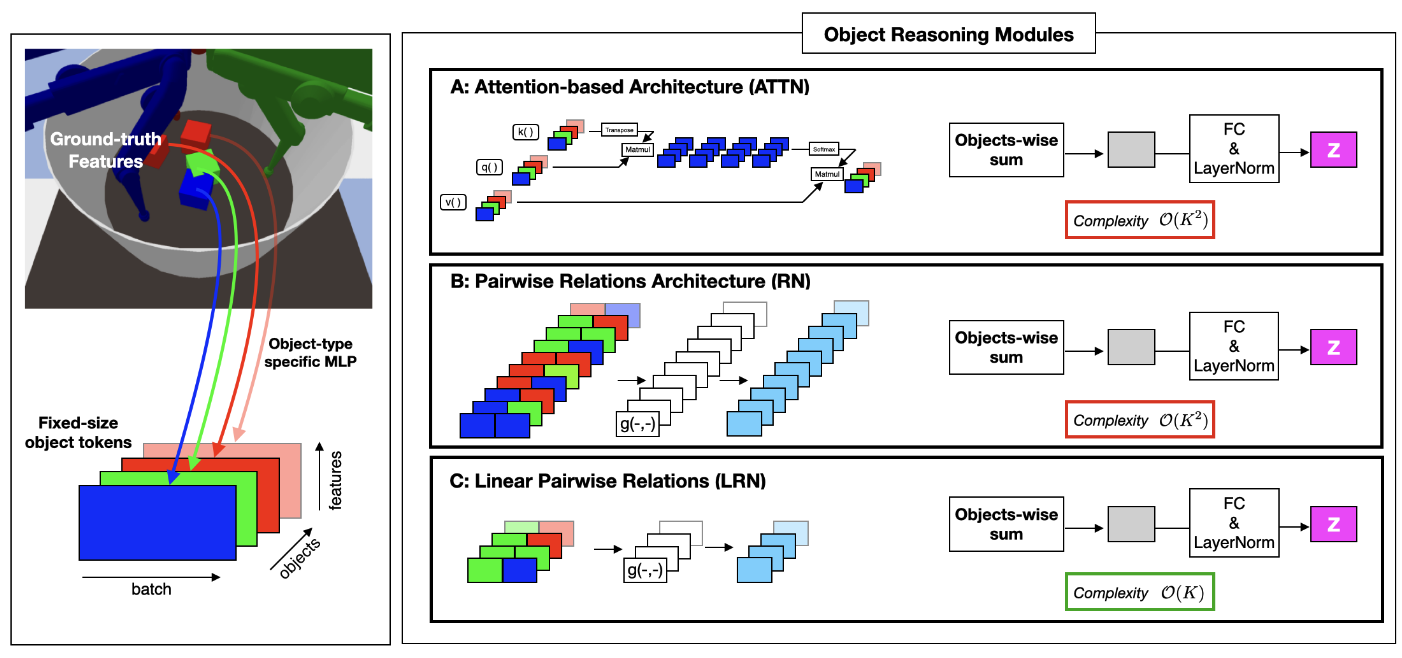}
\caption{Example of the reasoning of object relations in reinforcement learning \protect\cite{mambelli2022compositional}.
}\label{fig:exp_rl}
\end{figure}

\subsubsection{Multi-armed bandits}

In multi-armed bandits, compositional generalization refers to the ability of a model to learn and generalize to different reward structures and make correct generalizations about options in novel contexts. This ability is important for developing models that can make decisions in uncertain environments, such as online advertising or clinical trials, and maximize rewards over time. In \cite{saanum2021compositional} the authors examine the role of structure learning and compositionality in human reinforcement learning. They use a multi-armed bandit paradigm to investigate participants' ability to learn representations of different reward structures and combine them to make correct generalizations about options in novel contexts. The results showed that participants were able to transfer knowledge of simpler reward structures to make compositional generalizations about rewards in more complex contexts (e.g. Figure \ref{fig:exp_bandit}), allowing them to make more efficient decisions. The researchers also developed a computational model that was able to replicate these findings, and found it to be a better explanation of participant behavior than other models of decision-making and transfer learning. Overall, the study provides evidence that the ability to represent and generate structured relationships between options is crucial in human reward learning and decision-making.

\begin{figure}[H]
\centering
    \includegraphics[width=\linewidth]{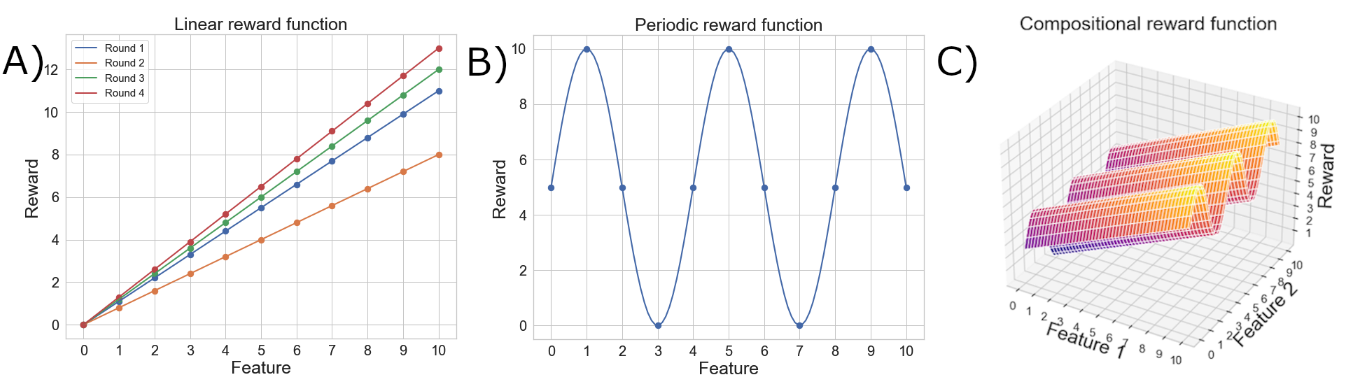}
\caption{Example of compositional vs. non-compositional rewards in bandits \protect\cite{saanum2021compositional}.
}\label{fig:exp_bandit}
\end{figure}

\subsection{Semantic parsing}
In semantic parsing, compositional generalization refers to the ability of a model to understand and generate structured predictions from new combinations of input sequences or more complex test structures. This ability is important for developing models that can understand natural language commands and questions and generate structured representations for a wide range of applications. In \cite{qiu2022evaluating}, the authors investigate whether increasing the size of pre-trained language models can improve their performance on out-of-distribution compositional generalization tasks in semantic parsing (Figure \ref{fig:exp_semantic}). They evaluate different model scaling methods, including fine-tuning, prompt-tuning and in-context learning, and compare their scaling curves. They find that fine-tuning generally has a flat or negative scaling curve, while in-context learning has a positive scaling curve, but is outperformed by smaller fine-tuned models. They also observe that prompt-tuning can outperform fine-tuning, and suggest that it has more potential for improving compositional generalization. Additionally, the researchers identify various error trends that vary with model scale, such as larger models being better at modeling the syntax of the output space, but also more prone to certain types of overfitting. Overall, the study highlights limitations of current techniques for leveraging model scale for compositional generalization and suggests promising directions for future research.

\begin{figure}[H]
\centering
    \includegraphics[width=\linewidth]{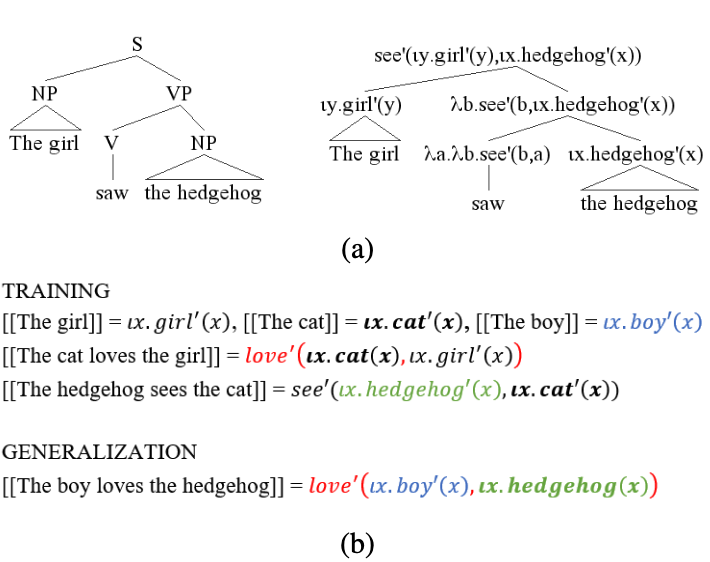}
\caption{Example of semantic compositional generalization (e.g. the meaning of the sentence is compositionally built up from the meanings of its parts) \protect\cite{kim2020cogs}.
}\label{fig:exp_semantic}
\end{figure}

\subsection{Image captioning}

In image captioning, compositional generalization refers to the ability of a model to understand and generate natural language descriptions of new images and image scenes that it has not encountered during training (e.g. Figure \ref{fig:exp_img}). This ability is important for developing models that can understand and describe images in a wide range of contexts, such as visual search, image retrieval, and autonomous systems.  In \cite{nikolaus2019compositional}, the focus is on compositional generalization in image captioning which measures a model's ability to compose new, unseen combinations of concepts when describing images. It is found that current state-of-the-art models do not perform well in this aspect. To address this, a multi-task model is proposed that combines caption generation and image-sentence ranking. The model uses a decoding mechanism that re-ranks the captions based on their similarity to the image. This approach significantly improves the model's ability to generalize to unseen combinations of concepts compared to current state-of-the-art models.

\begin{figure}[tbh]
\centering
    \includegraphics[width=\linewidth]{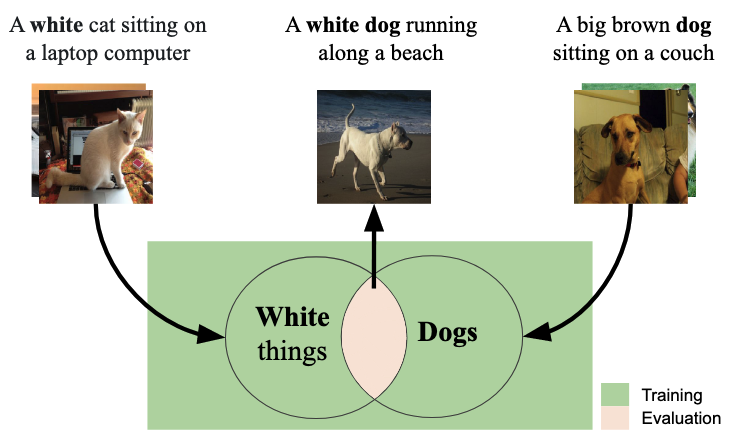}
\caption{Example of the compositional generalization in image captioning to unseen
combinations of adjectives, nouns, and verbs \protect\cite{nikolaus2019compositional}.
}\label{fig:exp_img}
\end{figure}

\subsection{Question answering}

In question answering, compositional generalization refers to the ability of a model to understand and generate answers to new combinations of questions and question structures that it has not encountered during training. This ability is important for developing models that can understand natural language questions and generate accurate answers for a wide range of applications, such as customer service, knowledge bases, and question-answering systems. In \cite{gai2021grounded}, the authors on improving the ability of question answering models to generalize to new and more complex combinations of training patterns. The current end-to-end models tend to lose important input syntax context when presented with novel compositions. In order to address this issue, the researchers propose a new method called Grounded Graph Decoding, which uses an attention mechanism to ground structured predictions, allowing the model to retain syntax information from the input. This method is able to learn a group-invariant representation without making assumptions about the target domain, and was found to significantly outperform existing state-of-the-art models on the Compositional Freebase Questions task.

\subsection{Automatic translation}

In automatic translation, compositional generalization refers to the ability of a model to understand and translate new combinations of phrases, sentences, and paragraphs that it has not encountered during training. This ability is important for developing models that can understand and translate natural language in a wide range of contexts, such as machine translation, multilingual search, and dialogue systems. \cite{li2021compositional} investigates the ability of modern neural machine translation (NMT) models to generalize well to new, unseen combinations of sentence components (Figure \ref{fig:exp_nmt}). They do this by creating a benchmark dataset called CoGnition, which includes a large number of clean and consistent sentence pairs. By analyzing the performance of NMT models on this dataset, the authors reveal that these models struggle with compositional generalization, despite performing well on standard metrics.

\begin{figure}[tbh]
\centering
    \includegraphics[width=\linewidth]{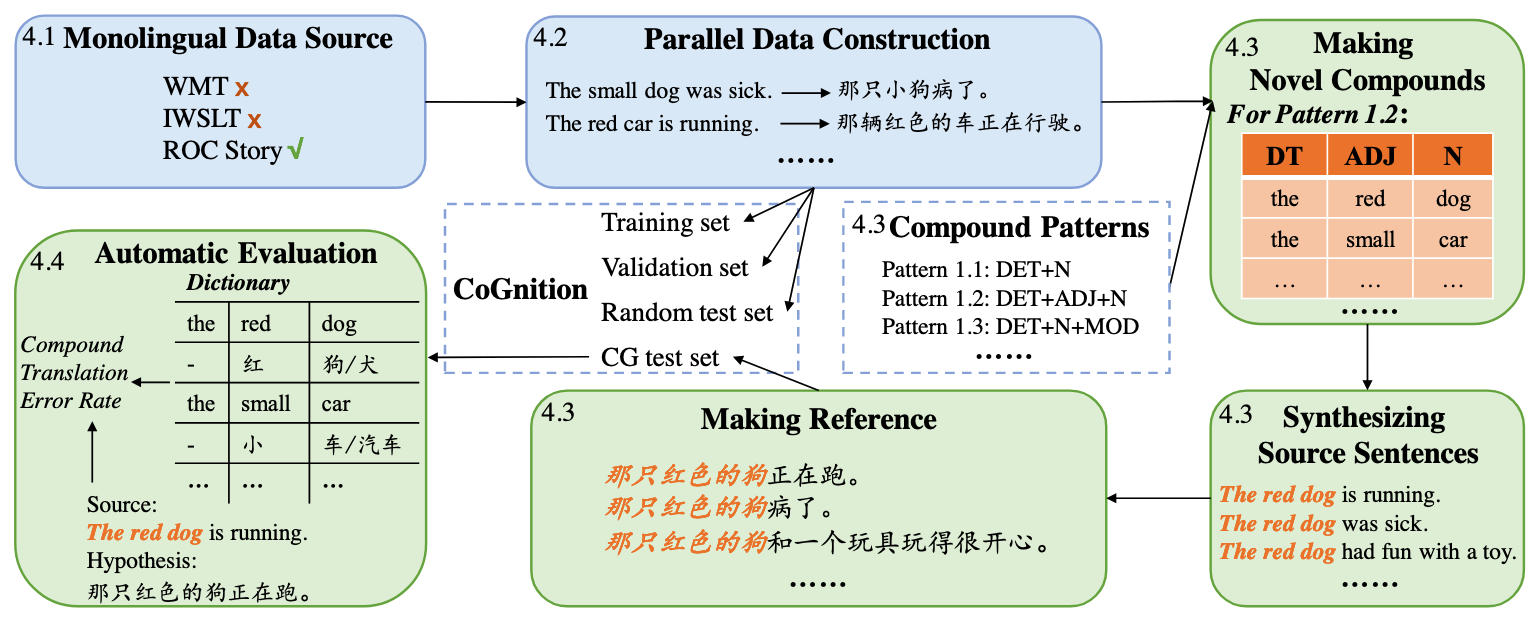}
\caption{Example of taking compositional generalization into account to train NMT models \protect\cite{li2021compositional}.
}\label{fig:exp_nmt}
\end{figure}

\subsection{Recommendation systems}

In recommendation systems, compositional generalization refers to the ability of a model to understand and predict user preferences based on new combinations of users, items, and interactions that it has not encountered during training. This ability is important for developing models that can make personalized recommendations in a wide range of contexts, such as e-commerce, online advertising, and social networks. In \cite{hang2022lightweight}, a new method called Lightweight Compositional Embedding (LCE) is proposed to improve the performance of graph-based recommendation systems in a streaming setting. Traditional recommendation systems rely on a static setting where all information about users and items is available upfront, but this is not feasible for real-world applications where data is constantly changing. LCE addresses this issue by only learning explicit embeddings for a subset of nodes and representing the others implicitly through a composition function based on their interactions in the graph. This allows for more efficient and effective updates to model predictions. The proposed method is evaluated on three large-scale datasets and shown to outperform other graph-based models, achieving nearly optimal performance with fewer parameters. These compositional generalization techniques can be useful in other domain-specific recommendation systems (e.g. healthcare recommendation \cite{lin2022supervisorbot}) where explicit embeddings and personalized recommendations are important to provide interpretable insights, efficient training and real time deployment.

\begin{figure}[tbh]
\centering
    \includegraphics[width=\linewidth]{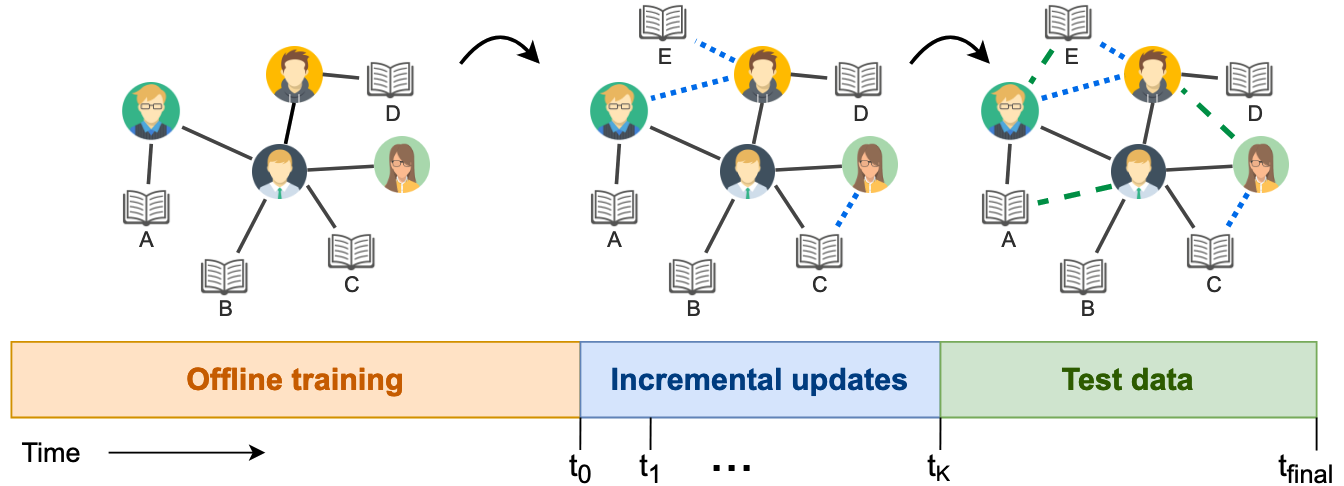}
\caption{Example of learning from the graphs of different users in recommendation systems considering their compositional generalization \protect\cite{hang2022lightweight}.
}\label{fig:exp_rs}
\end{figure}


 \subsection{Compositional Generalization in Real-Life Applications: Summary and Future Directions}
 
\begin{table}[h]
\scriptsize
\caption {Compositional Generalization in Real Life Applications}
\label{tab:Life} 
\begin{tabular}{|l|r|l|l|l|}
\hline
                            &     &     &       &     \\ 
                            & Systematicity    & Productivity  &   Rule Based      &  Neural  \\
                            &      &     &       &    \\
                            \hline
Healthcare        &  $\surd$ &                      &  $\surd$  &                                  \\ \hline
Finance           &   $\surd$ &                     &           &                                  \\ \hline
Dynamic pricing   &         &     $\surd$           &           &                                  \\ \hline
Recommendation system & $\surd$  &    $\surd$           & $\surd$   &     $\surd$                          \\ \hline
Maximization      & $\surd$  &                      &           &                                  \\ \hline
Dialogue system   &          &                      &   $\surd$  &                                  \\ \hline
Telecommunication  &  $\surd$ &                      &           &                                  \\ \hline
Anomaly detection           &   $\surd$ &                      &           &                                  \\ \hline
\end{tabular}
\end{table}

Table \ref{tab:Life} provides a summary of compositional generalization problem formulations used in various domain-specific applications. The choice of Compositional generalization model is often domain-specific. For example, we can observe that Systematicity is applied on all domain, and productivity is mainly used on dynamic pricing and recommendation system where the approach of generalization depend mainly on the similarity between features of the item recommended. The rule based approach is mainly used in more sensitive domain like healthcare, where the error in generalization is less tolerated. Right now we can observe that the neural approach for composition generalization is mainly used in recommendation system and could be expended to more domain. 

There are several trends in the field that are likely to shape the future of compositional generalization in AI. These include: (1) Advances in neural network architectures: The development of new neural network architectures that can better handle compositional data, such as graph neural networks and transformer models, will continue to advance the field. (2) Integration with other AI techniques: The integration of compositional generalization with other AI techniques, such as reinforcement learning and transfer learning, will lead to new approaches for building intelligent systems that can learn from experience and adapt to new situations. These insights from the compositional generalization, can also in turn, help create AI-augmented knowledge management systems that are modular and interpretable \cite{lin2022knowledge}.

The development of compositional generalization techniques can also be applied to new domains, such as finance and healthcare, or new subdomains of natural language processing and computer vision, which can further lead to new applications of AI. For instance, in the healthcare domain, compositional generalization can be used to analyze patient data and make predictions about potential diagnoses and treatment plans based on past patient experiences and outcomes. It can be used to analyze large amounts of data from clinical trials and generalize findings to new patient populations or treatments. 

Another example is the information retrieval. The authors in \cite{losada2017multi} argue that the iterative selection process of information retrieval can be naturally modeled as a contextual compositional generalization problem. The compositional generalization model leads to highly effective  methods for document adjudication. Under this compositional generalization allocation framework, they  propose seven new document adjudication methods, of which five are stationary methods and two are non-stationary methods. This comparative study includes existing methods designed for pooling-based evaluation and existing methods designed for metasearch. In mobile information retrieval, the authors in \cite{bouneffouf2013contextual} introduce an algorithm that tackles this dilemma in Context-Based Information Retrieval (CBIR) area. It is based on dynamic exploration/exploitation and it adaptively balances the two aspects by deciding which user's situation is most relevant for exploration or exploitation. Within a deliberately designed online framework, they conduct evaluations with mobile users. 

Compositional generalization is usually most suitable for the stationary settings, as opposed to specifically tailored dynamic solutions in non-stationary settings. Despite a consideration, there are still techniques to mitigate it in critical application domains. For instance, in healthcare applications, significant changes are not expected in the process of making the treatment decisions, i.e. no  transition in the state of the  the patient;  such transitions, if they occurred, would be better modeled using reinforcement learning rather than non-stationary compositional generalization. 

There are clearly other domains where the non-stationary compositional generalization is a more appropriate  setting, but it looks like this setting was  not yet been significantly investigated in healthcare domains. For example, anomaly detection, is a domain where non-stationary contextual compositional generalization could be used, since in this setting the anomaly could be adversarial, which means that any compositional generalization applied to this setting should have some kind of drift condition, in-order to adapt to new types of attacks. Another observation is that none of the existing work tried to develop an algorithm that could solve these different tasks at the same time, or apply the  knowledge obtained in one domain to another domain, thus opening a direction of research on {\em multitask} and {\em transfer learning} in compositional generalization setting. Furthermore, given an online nature of compositional generalization problem,  {\em continuous}, or {\em lifelong learning} would be a natural next step, adapting  the model learned in the previous tasks to the new one, while  still remembering how to perform earlier task, thus avoiding the  problem of ``catastrophic forgetting''.  One possible approach, for example, could include techniques such as pseudo-rehearsal, which involves constructing a pseudo-dataset by generating  random inputs, passing them through the original model and recording their output.
 
 \section{Conclusions}
\label{sec:Conclusion}
In this article, we reviewed some of the most notable recent work on applications of compositional generalization.in real-life domains and in automated machine learning. We summarized, in an organized way (Tables 1),  various existing applications, by types  settings used, and discussed the advantages of using compositional generalization techniques in each domain. We briefly outlines several important open problems and promising future extensions.

\bibliography{main}
\bibliographystyle{named}

\end{document}